\pdfoutput=1

\documentclass[11pt]{article}

\usepackage[]{EMNLP2022}

\usepackage{times}
\usepackage{latexsym}

\usepackage[T1]{fontenc}

\usepackage{array}
\usepackage{enumitem}
\usepackage{xcolor}
\usepackage{colortbl}
\usepackage{hhline}
\usepackage{CJKutf8}
\usepackage{booktabs}
\usepackage{multicol}
\usepackage{blindtext}
\usepackage{amssymb}
\usepackage{multirow}
\usepackage{adjustbox}
\usepackage{soul}
\usepackage{subcaption}
\usepackage[textsize=scriptsize]{todonotes}
\usepackage{pgfplots}
\usepgfplotslibrary{groupplots,dateplot}
\usetikzlibrary{patterns,shapes.arrows}
\pgfplotsset{compat=newest}

\definecolor{bblue}{HTML}{4F81BD}
\definecolor{oorange}{HTML}{F4C842}
\definecolor{rred}{HTML}{C0504D}
\definecolor{ggreen}{HTML}{9BBB59}
\definecolor{ppurple}{HTML}{9F4C7C}
\definecolor{darkgreen}{HTML}{228B22}
\definecolor{cred}{HTML}{D81B60}
\definecolor{cblue}{HTML}{1E88E5}
\definecolor{cyellow}{HTML}{FFC107}
\definecolor{nred}{HTML}{e41a1c}
\definecolor{nblue}{HTML}{377eb8}
\definecolor{ngreen}{HTML}{4daf4a}
\definecolor{lblue}{HTML}{6C8EBF}

\usepackage[utf8]{inputenc}
\usepackage{microtype}
\usepackage{xcolor}
\usepackage{booktabs}

\usepackage{inconsolata}

\usepackage{xspace}
\usepackage{graphicx} 
\graphicspath{ {./figs/} }
\title{Beyond Counting Datasets: \\A Survey of Multilingual Dataset Construction and Necessary Resources}

\newcommand\uw{$^\diamondsuit$}

\newcommand\ut{$^\clubsuit$}
\newcommand\uwm{$^\heartsuit$}

\usepackage{xcolor}

\author{Xinyan Velocity Yu\uw$^*$ \space\space\space Akari Asai\uw$^*$ \space\space\space Trina Chatterjee\ut\\ \space\space\space \textbf{Junjie Hu}\uwm  \space\space\space \space\space\space \textbf{Eunsol Choi}\ut  \\
\uw University of Washington, 
\uwm The University of Wisconsin-Madison, \ut The University of Texas at Austin \\
\small{\texttt{\{xyu530,akari\}@cs.washington.edu}},\\\small{\texttt{junjie.hu@wisc.edu}}, \small{\texttt{\{tchatter,eunsol\}@utexas.edu}}\\
}
\date{}

\begin{document}
\maketitle
\def\thefootnote{*}\footnotetext{Equal contributions.}
\begin{abstract}

While the NLP community is generally aware of resource disparities among languages, we lack research that quantifies the extent and types of such disparity. Prior surveys estimating the availability of resources based on the number of datasets can be misleading as dataset quality varies: many datasets are automatically induced or translated from English data. 
To provide a more comprehensive picture of language resources, we examine the characteristics of 156 publicly available NLP datasets. We manually annotate \textit{how} they are created, including input text and label sources and tools used to build them, and \textit{what} they study, tasks they address and motivations for their creation. 
After quantifying the {qualitative} NLP resource gap across languages, we discuss how to improve data collection in low-resource languages. We survey language-proficient NLP researchers and crowd workers per language, finding that their estimated availability correlates with dataset availability. Through crowdsourcing experiments, we identify strategies for collecting high-quality multilingual data on the Mechanical Turk platform. We conclude by making macro and micro-level suggestions to the NLP community and individual researchers for future multilingual data development.

\end{abstract}

\section{Introduction}
Datasets play fundamental roles in advancing language technologies~\cite{PAULLADA2021100336}. However, large disparities exist among languages in terms of the scale of existing datasets~\cite{kreutzer-etal-2022-quality,joshi-etal-2020-state} and the resulting task performance~\cite{blasi-etal-2022-systematic}. 
Multilingual resources also harbor unique annotation artifacts, such as translationese~\cite{clark-etal-2020-tydi,artetxe-etal-2020-translation}. Understanding dataset construction processes can help explain the true landscape of multilingual NLP datasets. 

Despite low-level awareness among multilingual researchers about disparities in linguistic resources, scant research to date has examined the \textit{quality} of labeled data in non-English languages. Existing surveys such as \citet{joshi-etal-2020-state} have studied the scale of available resources per language, without exploring how such datasets are developed, and often simply count number of datasets. 

\begin{figure}[t]
\resizebox{1.02\columnwidth}{!}{\includegraphics{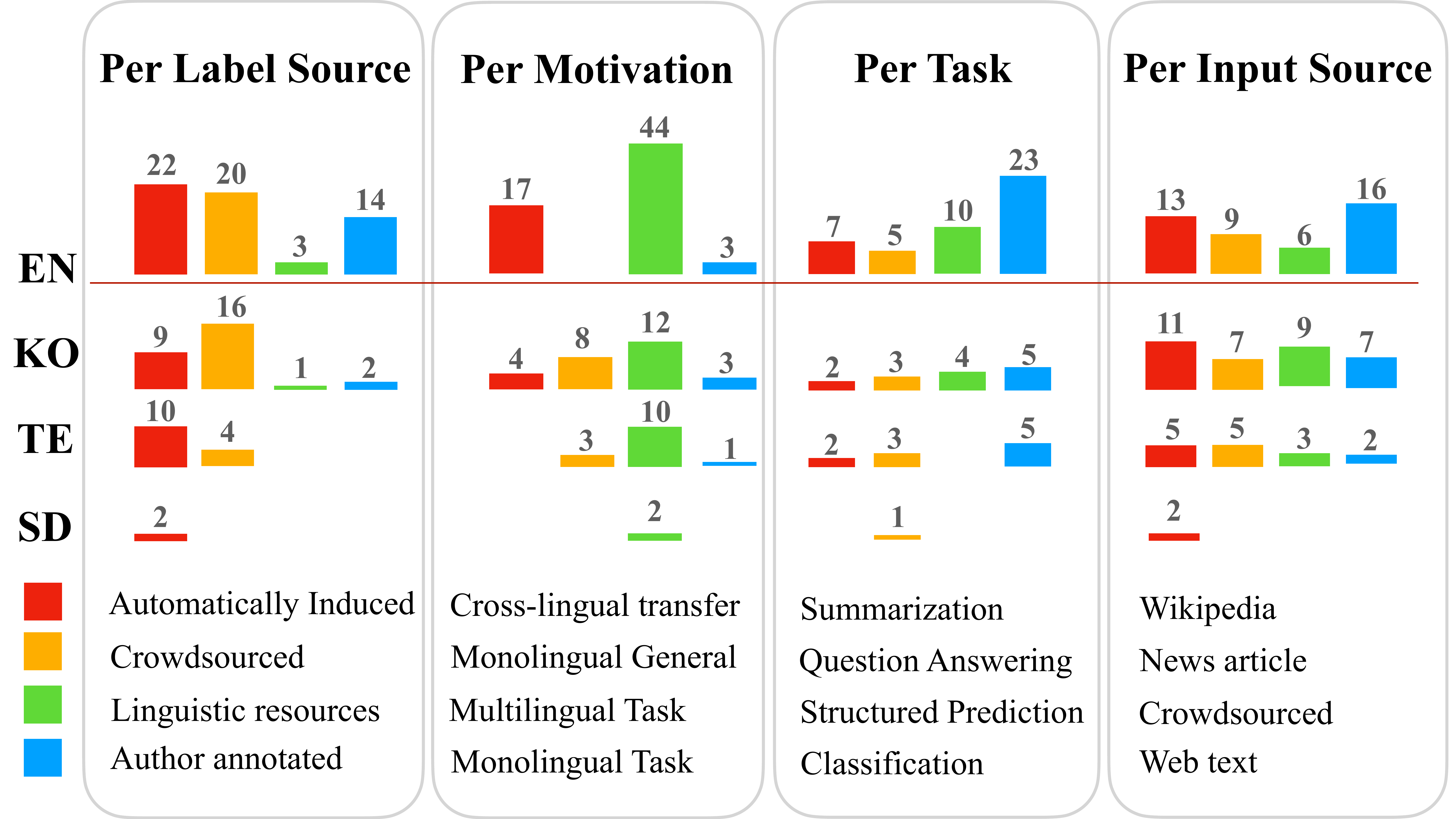}}\vspace{-.1cm}
\caption{Fine-grained, qualitative comparison of dataset availability in four languages: English (EN), Korean (KO), Telugu (TE) and Sindhi (SD). Low-resource languages often only feature automatically induced labels and lack diverse input sources (frequently limited to Wikipedia text) 
as a part of parallel multilingual evaluation datasets. }
\label{fig:teaser}\vspace{-0.7em}
\end{figure}

We provide the first large-scale survey of multilingual dataset characteristics, examining in depth how and why these datasets are constructed and quantifying their availability. We study 156 datasets, each covering at least one non-English language, from the ACL anthology and HuggingFace~\cite{lhoest-etal-2021-datasets}. 
{We propose a new annotation scheme with 13 attributes, } including the task it addresses, its source of input text, its label collection process, its stated motivation, and tools used to create it, such as translation services (Section~\ref{sec:annotation_scheme}). 

Our annotation yields rich information about the status of multilingual datasets (Section~\ref{sec:survey_results}). As shown in Figure \ref{fig:teaser}, language coverage varies significantly across tasks, and label collection methods differ across languages and tasks. 
222 languages are covered by the 156 datasets, and, on average, 5.6 datasets per language are available. However, we posit that this number is misleading when interpreting the landscape of multilingual NLP datasets:  we observe the prevalence of automatically induced labels particularly in low resource languages; one-third of the datasets we surveyed use automatically induced labels, and 68\% of the languages have no manually annotated data.  
While Wikipedia and news texts are available for a wide range of languages, texts specifically written for the NLP task 
are not available for most languages. 
Furthermore, about 20\% of surveyed datasets involved translation in their creation, and these were used at a much higher rate in highly cited resources to study cross-lingual transfer. These types of quantitative and qualitative disparities persist in multilingual supervised datasets, motivating us to identify underlying bottlenecks to create multilingual datasets.

We investigate whether the resources required for NLP data collection affect the prevalence of multilingual datasets. These include the availability of crowdworkers and NLP researchers with sufficient language proficiency as well as raw input text (Section~\ref{sec:correlation}). These data collection resources all correlate with the number of datasets in each language, suggesting paths for a more equitable data landscape. To assess the paths for crowdsourcing in non-English languages on a popular crowdsourcing platform, we design controlled data collection experiments in six languages, quantifying the challenges even in relatively high-resource languages. 

We conclude our survey by relating research-derived suggestions to the NLP community and to individual researchers for multilingual data construction. 
We also provide concrete suggestions for crowdsourcing platforms and crowdsourcing quality control tips, recommended translation services, and publication venues. We host our survey at \url{https://multilingual-dataset-survey.github.io}, permitting readers to review multilingual resources and submit their datasets following our schema.
\section{Related Work}
\citet{kreutzer-etal-2022-quality} study the quality of raw text data available for researchers, such as ParaCrawl~\cite{espla-etal-2019-paracrawl}, Wikimatrix~\cite{schwenk-etal-2021-wikimatrix}, and mC4~\cite{xue-etal-2021-mt5}, by manually evaluating 100 sample texts for each language. The data quality for low-resource languages is poor, and parallel sentences are often misaligned~\cite{koehn-etal-2019-findings}. 
\citet{blasi-etal-2022-systematic} focus on the models' performance on diverse tasks, suggesting that economic powers of the languages' users drives NLP technology development, {while we focus on analyzing the quality of existing multilingual datasets.} A recent position paper~\cite{hershcovich-etal-2022-challenges} discusses the representation of various cultures in NLP datasets. \citet{bender-friedman-2018-data} argue that providing details about data can help alleviate issues related to exclusion and bias in language technology. Our schema includes additional information such as text sources.  Most closely related to our work, \citet{joshi-etal-2020-state} study the availability of Linguistic Data Consortium (LDC) Catalog datasets and their relationship to certain factors (e.g., the number of Wikipedia articles). 
To our knowledge, our work is the first large-scale study (on 156 datasets beyond LDC catalog) of the curation process of labeled, open-sourced multilingual datasets and relevant external factors (e.g., the availability of crowdsourcing and language-proficient researchers).  
Several work study the quality and availability of crowdsourced workers to gather translation data on MTurk~\cite{callison-burch-2009-fast,bloodgood-callison-burch-2010-using,pavlick-etal-2014-language}. Our MTurk experiment examines qualification control methods for multilingual crowdsourcing.

\begin{table*}
\footnotesize
    \centering
        \begin{tabular}{p{2cm}|p{4cm}|p{8.5cm}}\toprule
        Aspect & Descriptions & Categories\\ \midrule
        Language$^\blacklozenge$ & Target languages & \{ISO 639-1 Language Code, not mentioned\}  \\
        \rowcolor[gray]{0.90} Task Type$^\blacklozenge$ & {Ten coarse-grained NLP task type that the dataset addresses} & \{classification (sentiment analyis), classification (sentence pair), classification (other), QA (w/ retrieval), QA (machine reading), structured prediction, sequence tagging, generation (summarization), generation (other), other\}  \\
        Dataset size & {Avg. \# of data in each language} & \{ $< 100$, $100\sim 1000$, $1000\sim10k$, $> 10k$ \}  \\ \midrule
         Creator$^\blacklozenge$ & Who led dataset creation &\{industry, individual researchers, university\}  \\
         Pub. Venue & Venue of paper publication  & \{*CL, LREC, *ACL Workshop, Findings, NeurIPS Datasets and Benchmarks Track, arXiv, N/A\} \\
         Pub.  Year & {Year of paper publication } &   Year of publication between 2008 - 2021 \\
         \rowcolor[gray]{0.90} Motivation & {Motivation for dataset creation} &  \{cross-lingual transfer, single task (multilingual) w/ ML training, single task (single lang), multi-task (single lang)\} \\ \midrule
     \rowcolor[gray]{0.90}Input text ($x$) Source$^\blacklozenge$ &Where input text is from  & \{annotated (authors, linguists), commercial sources, crowdsourced, curated linguistic resources (wordnet, etc.), curated source (exams, scientific papers, etc.), media, template-based, web, Wikipedia\} \\
         \rowcolor[gray]{0.90} $x$ Language & {Languages where $x$ is collected} & \{English, its own language, both, other language, not mentioned\} \\
         \rowcolor[gray]{0.90} Label ($y$) Collection$^\blacklozenge$ & How label is collected  & \{annotated (authors, linguists), automatically induced, crowdsourced, curated linguistic resources, not mentioned\}\\
        \rowcolor[gray]{0.90} $y$ Language & {Languages where $y$ is collected}  & \{English, its own language, not mentioned\} \\
        Reuse & Reusing released datasets? &  \{Yes-Eng, Yes-other-lang, Yes-Eng\&other-lang, No\} \\ 
         \rowcolor[gray]{0.90} Translation$^\blacklozenge$ & Type of translation used during data collection &\{automatic translation, human (author), human (non-author), no translation, unclear\}  \\
         \bottomrule
    \end{tabular}
    \vspace{-0.5em}
    \sethlcolor{lightgray}
    \caption{Annotation scheme. Each row represents an \textbf{aspect} that we label with \textbf{categories}. $\blacklozenge$ marks aspects on which a dataset can be assigned more than one categories. The \hl{highlight} marks an aspect where no prior survey has analyzed. {Input text ($x$) refers to the source of the text of the dataset, and label ($y$) refers to the target that the model should generate. For example, for a news summarization dataset, the label is the news summary and the input text source can be news articles from media.} }
    \vspace{-0.5em}
    \label{tab:annot_scheme}
\end{table*}
\section{Survey Scope, Scheme and Process}
\label{sec:annotation_scheme}
In the following section, we first identify the 156 open-sourced NLP datasets {published in NLP venues} that contain at least one non-English language. We then describe our proposed annotation scheme with 13 attributes, seven of which are analyzed for the first time (Table~\ref{tab:annot_scheme}).
\paragraph{{Scope.}} 
The scope of our work includes {\it labeled} datasets, where a system is expected to generate a label (output) $y$ given input text $x$. The output label is not limited to a single categorical label and includes generated text. We filter out (1) unlabeled data,  including parallel corpora (studied in \citealt{kreutzer-etal-2022-quality}); (2) language identification datasets;  (3) machine translation (MT) datasets (inherently cross-lingual); and (4) multi-modal datasets (other modalities can be language agnostic)

We compiled dataset lists in December 2021 sourced from: (1) the ACL paper anthology identified through a keyword search, and (2) the open-source Hugging Face Datasets after filtering excluded tasks and English-only datasets. {Our preliminary study shows that the combination of HuggingFace and ACL anthology keyword search gives reasonable coverage of existing multilingual NLP dataset papers: Most NLP dataset papers have been published in ACL-related conferences (e.g., LREC, *CL workshops, *CL) and listed in ACL anthology. HuggingFace datasets\footnote{\url{https://huggingface.co/docs/datasets/index}} are widely used in the NLP community and many recently-published papers make their datasets available there.}
We searched the ACL anthology using case-insensitive keyword matches by first selecting papers that includes specific keywords in their titles, and then filtering out the papers mentioning excluded tasks in their titles and abstracts: 
\vspace{-.2em}
\begin{itemize}[leftmargin=1.3em]
    \setlength\itemsep{-0.1em} 
    \item \textbf{Included keywords:} multilingual, cross-lingual, dataset, annotation, labelled, benchmark. 
    \item \textbf{Excluded keywords:} machine translation, language identification, vision, topic model, induction, speech, multimodal.
\end{itemize}

We selected these keywords to focus our study on labeled multilingual NLP \textit{text-only} datasets. The preceding steps helped us select 151 papers of the 73,384 papers in the ACL anthology, and we subsequently 
filtered out papers due to unavailability of datasets. 
For multi-task datasets for a specific language (such as KLUE;~\citealt{park2021klue}), we decomposed the unified benchmark into each sub-task and included each sub-task as a unique dataset. {For example, in the KLUE benchmark, there are eight tasks in total, and we annotate each sub-task dataset independently. }
Standardized multi-task test suites that directly reuse existing resources such as XTREME~\cite{hu2020xtreme}, {or dataset / sub-task dataset that directly reuses a previously published dataset with no significant modification to any of the schemes} 
were not included, {and we only include the original dataset.} Our final survey includes 156 datasets from 112 papers.
  \begin{table*}
\footnotesize
    \centering
    \addtolength{\tabcolsep}{-1.5pt}    

    \begin{tabular}{l|ccc|l}
    \toprule
    Category & Lang & Task  & Train & Main Goal (an example dataset)\\
    \midrule
    cross-lingual transfer & multi & single & & Evaluating across languages w/o training (e.g., XNLI;~\citealt{conneau2018xnli}) \\
multilingual task & multi & single & \checkmark & Improving a task across languages (e.g., TyDi QA;~\citealt{clark-etal-2020-tydi})\\
    monolingual task & single & single & \checkmark  & Improving a task in a single language (e.g., KQuAD;~\citealt{Lim2019KorQuAD10KQ}) \\
    monolingual general & single & multi & \checkmark & Improving multiple tasks in a single language (e.g., CLUE;~\citealt{xu-etal-2020-clue}) \\
        \bottomrule
    \end{tabular}\vspace{-0.4em}
     
\addtolength{\tabcolsep}{1.5pt}
    \caption{Motivation category explanations. The \textit{Train} column indicates presence/absence of a data split. We label datasets with training data as\textit{ cross-lingual transfer} when the original papers explicitly mention that as their primary goal rather than developing a system for the target task. }
    \label{tab:motivations_overview}
\end{table*}
\paragraph{Annotation Scheme.}
Table~\ref{tab:annot_scheme} describes the annotation scheme, consisting of \textbf{aspects} (e.g., task type) and their \textbf{categories} (e.g., summaries, sequence tagging). 
We cover three topics related to each dataset: (1) \textit{coverage}, i.e., what languages and tasks does it cover? (2) \textit{metadata}, i.e., why and who created the dataset? when was it created? (3) \textit{source}, i.e., how were the input and labels collected? We started with a set of categories for each aspect and updated it periodically. {Four (motivation, use/type of translation, input language, output language) out of our 13 annotation attributes focus on multilingual dataset collection. The rest of the scheme can also be useful to analyze the distribution of English datasets for future work.}

Many aspects are self-explanatory (e.g., language of the dataset), but the {\it motivation} aspect requires elaboration. 
While datasets are often repurposed after their introduction, 
dataset creators aim to address specific research questions at the creation time. We identified four types of motivation after manually reading the papers, including cross-lingual transfer, multilingual task, monolingual task, and monolingual general, which are summarized and explained in Table~\ref{tab:motivations_overview}. 
\paragraph{Annotation Process.}
{Three of the authors of this paper, each of whom have more than one year of NLP research experience and speaks more than one language, have manually annotated the datasets from December 2021 until March 2022. When a dataset fell within the boundaries of the pre-defined category set, each annotator tagged each borderline case, and all authors resolved the final category by discussion. }For multi-category aspects noted by $^\blacklozenge$ in Table~\ref{tab:annot_scheme}, we incremented \textit{the count of each}.



\section{Survey Results}
\label{sec:survey_results}
We summarize our findings, focusing on novel aspects. For each category of each aspect, we report the number of datasets belonging to that category and the unique languages they cover. We also highlight noteworthy correlations between multiple aspects, e.g., how label collection methods correlated with the type of tasks addressed. %
\begin{table}
\footnotesize
    \centering
    \begin{tabular}{llrr}
    \toprule
    \multicolumn{2}{c}{Task type} & \# Data & \# Langs\\
    \midrule
        Classification & sentiment  & 20 & 29\\
        & sentence pair & 17 & 25\\
        & other & 33 & 108\\ \midrule
        \multicolumn{2}{l}{Sequence tagging } & 16 & 194\\\midrule
        Generation & summarization & 10 & 95\\ 
        & other  & 9 & 12\\\midrule
        QA & machine reading & 20 & 47\\
        &  w/ retrieval  & 10 & 142\\\midrule
        \multicolumn{2}{l}{Structured prediction} & 19 & 33\\\midrule
        Other & & 3 & 13 \\ 
        \bottomrule
    \end{tabular}\vspace{-0.6em}
    \caption{The statistics of datasets by task types. 
}
    \label{tab:tasks}
\end{table}
\subsection{Coverage}
\paragraph{Task Types.}
We classify NLP tasks into five major categories: structured prediction, sequence tagging, generation, question answering (QA) and classification. For the last three, we provide sub-categories. 
Table~\ref{tab:tasks} shows per-task resources availability. Sequence tagging, summarization, and information retrieval have a wide coverage of languages. 
However, our analysis reveals that many datasets of these tasks were constructed with distant supervision. Classification task is common, while fewer resources are available for complex tasks such as structured prediction.\footnote{We interpret ``structured prediction'' coarsely, referring to tasks with a complex output space, such as parsing, coreference resolution, dialogue state tracking, and discourse analysis. } 
Limited resources exist for text generation, potentially because of the difficulty of quality control for generated texts. As an exception, summarization is covered broadly as they are often scraped from encyclopedic or news websites~\cite{hasan-etal-2021-xl}, although recent work shows such automatically created summarization datasets can be noisy~\cite{goyal2022news}. 

\begin{figure}
\centering 
  \includegraphics[width=\linewidth]{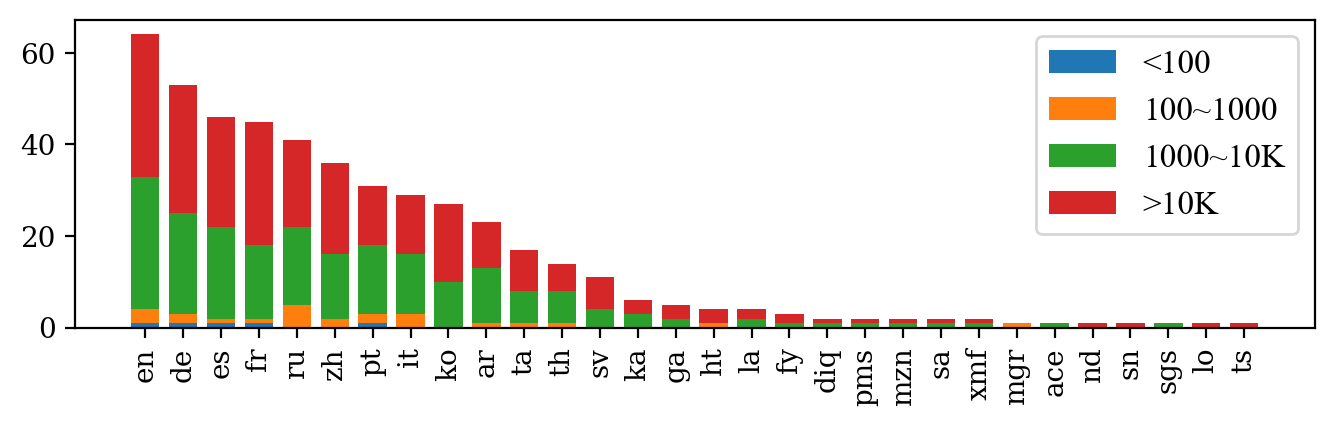}
 \caption{The number of datasets per data size bucket (the number of examples in the dataset) per language. 
  }
\label{fig:lang_vs_size}
\end{figure}

\paragraph{Dataset Size.} Each bar in Figure~\ref{fig:lang_vs_size} presents the number of datasets bucketed by data size for a specific language for the top 10 languages and 20 randomly sampled languages. 
Low-resource languages do not necessarily have smaller datasets, but they have fewer manually annotated datasets. 
Automatically induced datasets, where label $y$ is determined without human supervision, 
also tend to be larger: 35 out of 81 datasets that have more than 10K examples are automatically induced.

\subsection{Input and Label Collection}
\paragraph{Input Source: Collection Process for Input Texts ($x$).}
\begin{table}[t!]
    \centering
    \footnotesize
    \begin{tabular}{llrr}
    \toprule
 Source & Category & \# Data & \# Langs \\
    \midrule
        generated  & crowdworkers & 16 & 33  \\
       by & authors, linguists &2 & 9 \\
         & template & 2 & 11  \\\midrule
        collected  & web & 35 & 114\\
      from & social media/commerce  & 24 & 33  \\
        & Wikipedia & 32 & 184\\
        & media (news)  & 40 & 103\\\midrule
        curated   & linguistics & 15 & 32  \\
      source &  others (exams, etc)  & 25 & 59\\
        \bottomrule
    \end{tabular}
\vspace{-0.5em}
    \caption{Statistics on the source of input texts. }
    \label{tab:input_source}
\vspace{-0.5em}
\end{table}
We classify input text source into three high-level categories {(i.e., \textit{generated by} human, \textit{collected from} websites, and extracted from \textit{curated sources})} and break them down into nine fine-grained categories. Table~\ref{tab:input_source} shows the categories and annotation results. While we cover 222 languages, only 40 of them (18\%) 
have input text specifically written by humans for the task.\footnote{Due to overlap of language covered among datasets, there are 40 unique languages instead of 53 (from Table~\ref{tab:input_source}).}
The news is the most common source, used by 40 of 156 datasets, often in summarization or classification tasks, 
followed by web corpora\footnote{The ``web'' fine-grained category refers to the collection of sentences scraped or sampled at large-scale from the web.} and Wikipedia. Many languages have datasets derived only from Wikipedia text. 
Figure~\ref{fig:task_source_relationship} in appendix shows per-task input source distributions. 

\begin{table*}
    \centering
    \footnotesize
    \begin{tabular}{p{4cm}p{8cm}rr}
    \toprule
    Label source & Description &  \# Data & \# Langs  \\ \midrule
    annotated by authors or linguists & manual annotation by domain experts. & 37 & 43 \\
    crowdsourced & manual annotation by crowdworkers.& 63 & 56 \\
    automatically induced & automatically aligned or deduced from labeled or unlabeled data.&53& 210 \\
    linguistic data & derived from curated linguistic resources (e.g., WordNet). & 5 & 24\\ \midrule
    not mentioned & No details provided or inadequate documentation.&9 & 18\\
    \bottomrule
    \end{tabular}\vspace{-0.3em}
    \caption{Label collection method statistics. If dataset creation involved multiple methods (e.g., automatically induced and then manually verified by authors), they are counted for each dataset.}
    \label{tab:label_ontology}
\end{table*}
Overall, we observe that high-resource languages entertain a variety of input sources, while low-resource ones rely on fewer resources such as Wikipedia and news. 

\paragraph{Label Source: Collection Process for Labels ($y$).}


Table~\ref{tab:label_ontology} presents the statistics on how the output labels were collected, split into five categories: {annotated by authors or linguists}, crowdsourced, automatically induced, derived from linguistic resources, and not mentioned.
Label collection methods affects dataset quality. While manually annotated datasets can exhibit artifacts~\cite{gururangan-etal-2018-annotation,poliak-etal-2018-hypothesis}, they are often validated via inter-annotator agreement. 
In contrast, automatically induced datasets are often introduced without such kind of validation phases and tend to be noisy. For example, bullet points from news article are often considered to be the summary of the article, which can contain missing background information in the rest of the article~\cite{Kang2020ImprovedNL,goyal2022news}.
Fifty-three datasets had automatically induced labels, most commonly seen for summarization and classification tasks, and 95 used manual annotation. 
This further breakdowns to 27 datasets \textit{solely} annotated by domain experts and 56 datasets \textit{solely} annotated by crowdworkers.
We investigated annotator pools for non-English languages in Section \ref{sec:correlation}.

\vskip .2cm \noindent {\bf Label Source and Task Types.}
Figure~\ref{fig:task_relationship} presents label collection methods per task type. QA with retrieval (e.g., XQA;~\citealt{liu-etal-2019-xqa}) and generation tasks show a high proportion of automatically induced datasets. 
In contrast, structured prediction datasets were rarely automatically induced; they were more often annotated by authors or linguists. 
Crowdsourcing is commonly used to construct reading comprehension and classification datasets. 

\vskip .2cm \noindent {\bf Label Source and Language Diversity.}
Figure~\ref{fig:labels_top20} shows the distributions of the label data collection methods for the top 10 languages and for 20 sampled languages. 
In high-resource languages, a large number of datasets are labeled manually, 
where in low-resource languages, the percentage of automatically induced datasets increases, with 135 languages have only  automatically induced datasets. 
On (macro-)average, the 10 highest resource languages show {43.4}\% of their datatsets with only automatically induced labels; however, for all languages, {84.9}\% of the datasets use only automatically induced labels. Prior work often uses the total number of datasets in a target language as a proxy for resource availability of the language, which our analysis suggests is limited. 

\begin{figure}[t!]
\begin{tikzpicture}[scale=0.5]

\definecolor{crimson2143940}{RGB}{214,39,40}
\definecolor{darkgray176}{RGB}{176,176,176}
\definecolor{darkorange25512714}{RGB}{255,127,14}
\definecolor{forestgreen4416044}{RGB}{44,160,44}
\definecolor{lightgray204}{RGB}{204,204,204}
\definecolor{steelblue31119180}{RGB}{31,119,180}

\begin{axis}[
legend cell align={left},
legend columns=2,
legend style={
  fill opacity=1,
  draw opacity=1,
  text opacity=1,
  at={(0.5,1.2)},
  anchor=north,
  draw=lightgray204
},
tick align=outside,
tick pos=left,
x grid style={darkgray176},
xmin=0, xmax=36.75,
xtick style={color=black},
y grid style={darkgray176},
ymin=-0.5, ymax=9.5,
ytick style={color=black},
ytick={0,1,2,3,4,5,6,7,8,9},
yticklabels={
  Other,
  Structured prediction,
  QA (machine reading),
  QA (w/ retrieval),
  Generation (other),
  Generation (summarization),
  Sequence tagging,
  Classification (sentence pair),
  Classification (other),
  Classification (sentiment analysis)
}
]
\draw[draw=none,fill=steelblue31119180] (axis cs:0,-0.25) rectangle (axis cs:1,0.25);
\addlegendimage{ybar,ybar legend,draw=none,fill=steelblue31119180}
\addlegendentry{annotated (authors, linguists)}

\draw[draw=none,fill=steelblue31119180] (axis cs:0,0.75) rectangle (axis cs:10,1.25);
\draw[draw=none,fill=steelblue31119180] (axis cs:0,1.75) rectangle (axis cs:3,2.25);
\draw[draw=none,fill=steelblue31119180] (axis cs:0,2.75) rectangle (axis cs:0,3.25);
\draw[draw=none,fill=steelblue31119180] (axis cs:0,3.75) rectangle (axis cs:2,4.25);
\draw[draw=none,fill=steelblue31119180] (axis cs:0,4.75) rectangle (axis cs:2,5.25);
\draw[draw=none,fill=steelblue31119180] (axis cs:0,5.75) rectangle (axis cs:4,6.25);
\draw[draw=none,fill=steelblue31119180] (axis cs:0,6.75) rectangle (axis cs:4,7.25);
\draw[draw=none,fill=steelblue31119180] (axis cs:0,7.75) rectangle (axis cs:7,8.25);
\draw[draw=none,fill=steelblue31119180] (axis cs:0,8.75) rectangle (axis cs:2,9.25);
\draw[draw=none,fill=darkorange25512714] (axis cs:1,-0.25) rectangle (axis cs:3,0.25);
\addlegendimage{ybar,ybar legend,draw=none,fill=darkorange25512714}
\addlegendentry{automatically induced}

\draw[draw=none,fill=darkorange25512714] (axis cs:10,0.75) rectangle (axis cs:11,1.25);
\draw[draw=none,fill=darkorange25512714] (axis cs:3,1.75) rectangle (axis cs:9,2.25);
\draw[draw=none,fill=darkorange25512714] (axis cs:0,2.75) rectangle (axis cs:6,3.25);
\draw[draw=none,fill=darkorange25512714] (axis cs:2,3.75) rectangle (axis cs:7,4.25);
\draw[draw=none,fill=darkorange25512714] (axis cs:2,4.75) rectangle (axis cs:10,5.25);
\draw[draw=none,fill=darkorange25512714] (axis cs:4,5.75) rectangle (axis cs:7,6.25);
\draw[draw=none,fill=darkorange25512714] (axis cs:4,6.75) rectangle (axis cs:7,7.25);
\draw[draw=none,fill=darkorange25512714] (axis cs:7,7.75) rectangle (axis cs:20,8.25);
\draw[draw=none,fill=darkorange25512714] (axis cs:2,8.75) rectangle (axis cs:7,9.25);
\draw[draw=none,fill=forestgreen4416044] (axis cs:0,-0.25) rectangle (axis cs:0,0.25);
\addlegendimage{ybar,ybar legend,draw=none,fill=forestgreen4416044}
\addlegendentry{crowdsourced}

\draw[draw=none,fill=forestgreen4416044] (axis cs:11,0.75) rectangle (axis cs:16,1.25);
\draw[draw=none,fill=forestgreen4416044] (axis cs:9,1.75) rectangle (axis cs:22,2.25);
\draw[draw=none,fill=forestgreen4416044] (axis cs:6,2.75) rectangle (axis cs:9,3.25);
\draw[draw=none,fill=forestgreen4416044] (axis cs:0,3.75) rectangle (axis cs:0,4.25);
\draw[draw=none,fill=forestgreen4416044] (axis cs:0,4.75) rectangle (axis cs:0,5.25);
\draw[draw=none,fill=forestgreen4416044] (axis cs:7,5.75) rectangle (axis cs:14,6.25);
\draw[draw=none,fill=forestgreen4416044] (axis cs:7,6.75) rectangle (axis cs:17,7.25);
\draw[draw=none,fill=forestgreen4416044] (axis cs:20,7.75) rectangle (axis cs:35,8.25);
\draw[draw=none,fill=forestgreen4416044] (axis cs:7,8.75) rectangle (axis cs:17,9.25);
\draw[draw=none,fill=crimson2143940] (axis cs:0,-0.25) rectangle (axis cs:0,0.25);
\addlegendimage{ybar,ybar legend,draw=none,fill=crimson2143940}
\addlegendentry{linguistic}

\draw[draw=none,fill=crimson2143940] (axis cs:16,0.75) rectangle (axis cs:18,1.25);
\draw[draw=none,fill=crimson2143940] (axis cs:0,1.75) rectangle (axis cs:0,2.25);
\draw[draw=none,fill=crimson2143940] (axis cs:9,2.75) rectangle (axis cs:10,3.25);
\draw[draw=none,fill=crimson2143940] (axis cs:0,3.75) rectangle (axis cs:0,4.25);
\draw[draw=none,fill=crimson2143940] (axis cs:0,4.75) rectangle (axis cs:0,5.25);
\draw[draw=none,fill=crimson2143940] (axis cs:14,5.75) rectangle (axis cs:15,6.25);
\draw[draw=none,fill=crimson2143940] (axis cs:17,6.75) rectangle (axis cs:18,7.25);
\draw[draw=none,fill=crimson2143940] (axis cs:0,7.75) rectangle (axis cs:0,8.25);
\draw[draw=none,fill=crimson2143940] (axis cs:0,8.75) rectangle (axis cs:0,9.25);
\end{axis}

\end{tikzpicture}
\caption{The distribution over label collection methods per task type. The size of bar for each collection method represents the number of datasets of that task type. \label{fig:task_relationship}}
\end{figure}
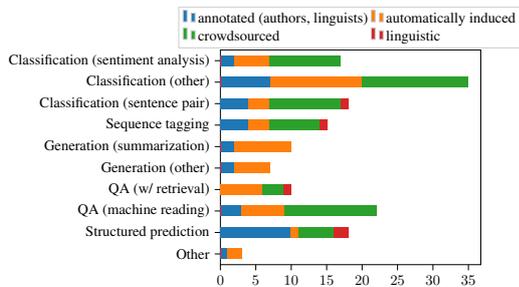
\begin{figure}[t!]
    \centering
    \includegraphics[width=\linewidth, keepaspectratio]{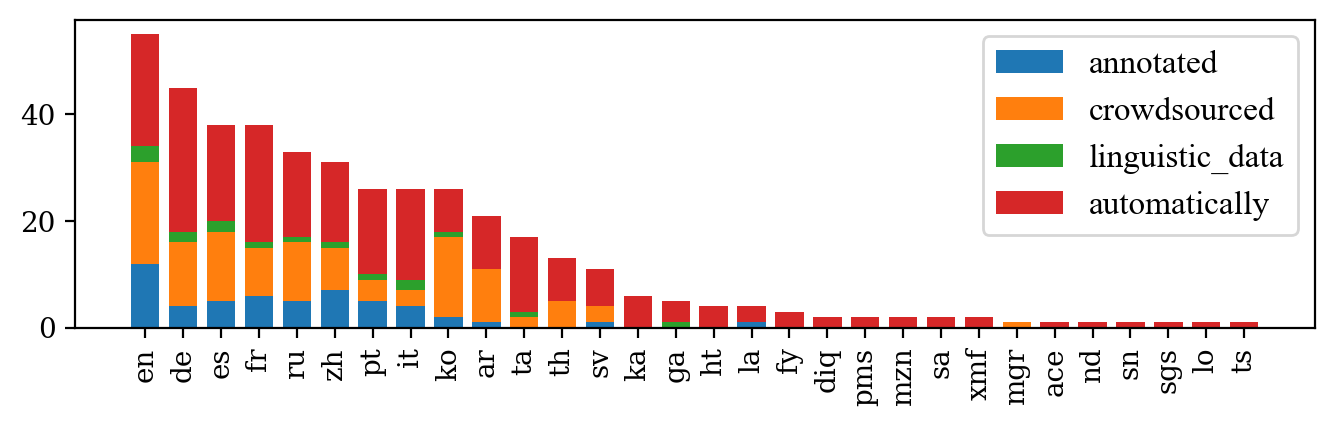}
    \caption{Label source per language for the top 10 and the  {20} sampled languages below top 10.}
    \label{fig:labels_top20}
\end{figure}


\subsection{Translations}
Many datasets are created by translating existing high-resource language datasets into target languages, which allows the creation of parallel data across many languages and removes reliance on the limited number of language-proficient annotators.
\begin{table}[t!]
    \footnotesize
    \centering
    \begin{tabular}{llrr}
    \toprule
    \multicolumn{2}{l}{Translation involved} & \# Data & \# Langs \\
    \midrule
        Yes   &automatic  & 12 & 178 \\
        & human (author) & 2 & 3\\
        & human (non-author)& 15 & 37 \\
         & unclear & 7 & 8 \\
         \midrule
       No &  & 123 & 185 \\
        \bottomrule
    \end{tabular}
    \vspace{-0.5em}    \caption{Statistics on translation involved. }
    \label{tab:translation}
    \vspace{-0.5em}
\end{table}
Table~\ref{tab:translation} reports the number of datasets and language covered by different translation methods. 
Thirty-three datasets used some translation during the creation process, compared to 123 that did not.  
Quality issues can arise when using automatic translation for dataset creation; these include quality degradation for long sentences and translation artifacts~\cite{lembersky-etal-2011-language, eetemadi-toutanova-2014-asymmetric, koehn-knowles-2017-six, artetxe-etal-2020-translation}; \citet{clark-etal-2020-tydi} suggests that besides translation artifacts, translation-based approaches can result in data that does not reflect native speakers' interests. Despite these problems, automatic translation was still used for 12 datasets. 

\paragraph{Input and Label Derivation via Translation.}
Translating English data into the target languages are used in 33 datasets, but most datasets collected data in its original language. 
Yet, many recent and highly cited datasets for cross-lingual transfer evaluation~\cite{artetxe-etal-2020-cross,conneau2018xnli} are created with translation-based approaches, which we discuss in detail below. 

\subsection{Motivation for Dataset Creation}
Table \ref{tab:motiv_vs_trans} summarizes statistics on the motivation aspect, with a breakdown for the number of datasets for each motivation with and without translation.
The most frequent driver for dataset creation was to cover multiple languages for a single task (62 datasets, covering 217 languages), often for downstream tasks with high economic demands, such as QA or summarization~\cite{blasi-etal-2022-systematic}. 
There were 16 languages (e.g., Chinese, Arabic) that had their own benchmark suites labeled as the monolingual general model category, which seemed to align with the availability of language-proficient NLP researchers. We discuss the relationship between dataset availability and the number of language-proficient researchers in Section~\ref{sec:correlation}.

\paragraph{Motivation and Translation Used.} 
The datasets studying cross-lingual transfer used translation at a much higher rate (61.9\%) than datasets with other motivations. Monolingual task datasets rarely used translation (8.5\%). 

\paragraph{Motivation and Affiliation.} We also find that  (1) industry researchers focused on cross-lingual transfer (12 of 20 papers), while academic researchers focused more on single task-oriented (either multi- or monolingual task) benchmarks (67 of {80} papers), and 
(2) MRC and QA 
had a higher proportion of task-oriented datasets than other tasks, potentially because they are close to downstream products. 

\begin{table}[t!]
\footnotesize
\centering
\begin{tabular}{p{3cm}|rr|rr}
\toprule
     \multirow{2}{*}{Motivation} & \multicolumn{2}{c|}{Translation}&\# Data & \# Lang\\
     & No & Yes &  \\\midrule
     cross-lingual transfer  & 8 & 13 & 21 & 48\\
     multilingual task & 52 & 11 & 62 & 217\\
     monolingual task & 32 & 3 & 35 & 25\\
     monolingual general & 31 & 6 &37 & 16 \\
     \bottomrule
\end{tabular}\vspace{-0.6em}
\caption{Motivations for dataset creation and their reliance on external translations.}
\vspace{-0.5em}
\label{tab:motiv_vs_trans}
\end{table}


\section{What's Needed to Develop NLP Datasets for Global Languages?}
\label{sec:correlation}
{We study the building blocks to create multilingual datasets:} (1) NLP researchers who speak the target language, and (2) ways to collect labels in the target language, including hiring crowdsourced workers. We computed the Pearson correlation coefficient ($\rho$) between the number of surveyed datasets and the proxy for each building block.\footnote{Appendix \ref{sec:input-text-avail} studies the availability of unlabeled text data in a target language and the number of surveyed datasets, finding a positive correlation, as in prior work~\cite{joshi-etal-2020-state} which surveyed LDC datasets.}

\begin{figure}[t!]
\centering
\begin{subfigure}[t]{.5\linewidth}
  \centering
  \includegraphics[width=\textwidth,keepaspectratio]{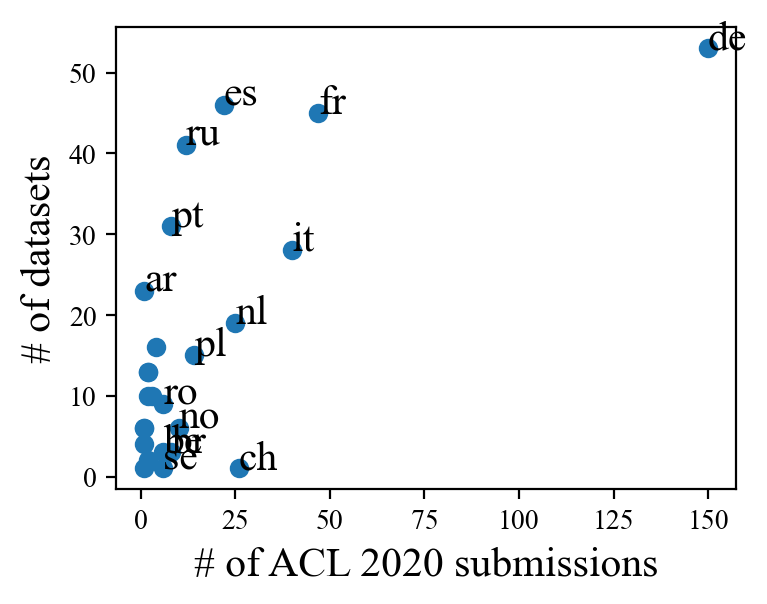}
  \caption{The \# of the ACL 2020 submissions from the countries where the languages are spoken.}
  \label{fig:acl2020}
\end{subfigure}%
\begin{subfigure}[t]{.5\linewidth}
  \centering
   \includegraphics[width=\textwidth,keepaspectratio]{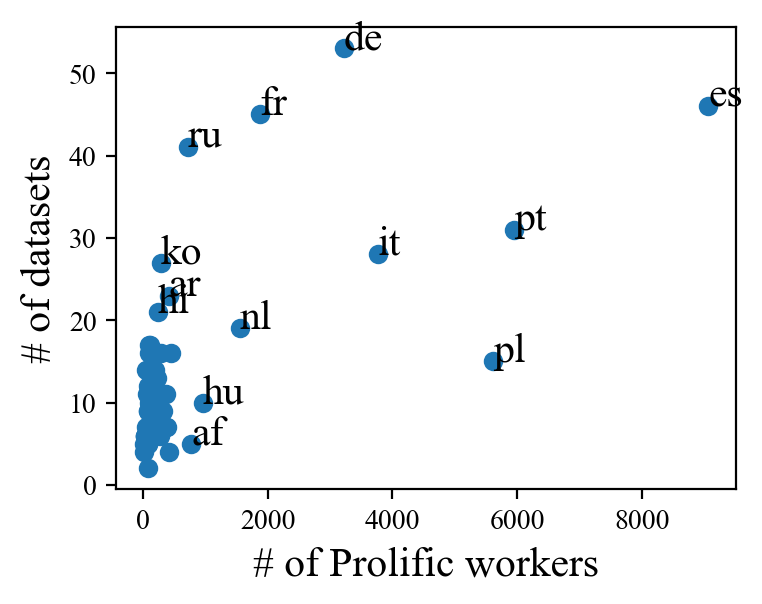}
\captionsetup{width=0.8\textwidth}
  \caption{The number of active crowdworkers on Prolific.}
  \label{fig:prolific_data_num}
\end{subfigure}
\caption{The number of non-English datasets and the factors that correlates. Each dot represents a language.}
\vspace{-1em}
\label{fig:bottlenecks}
\end{figure}
\paragraph{Availability of NLP Researchers.} 
Collecting data in a language without the availability of someone who understands the task and language to patrol data collection is challenging. We approximate the number of NLP researchers with language proficiency by the number of submissions to ACL 2020 from released general conference statistics.\footnote{\url{https://acl2020.org/blog/general-conference-statistics/}} We heuristically map country names to a set of languages commonly spoken in those countries. We use the number of submissions as a proxy to reflect research activities. 
Figure~\ref{fig:acl2020} shows a scatter plot for all surveyed languages except two highly dominant languages (i.e., English and Chinese). Its $x$-axis is the number of ACL 2020 submissions, and its $y$-axis is the number of labeled datasets, which are correlated with $\rho = 0.57$.
The correlation between the number of manually annotated datasets and the number of researchers was even higher ($\rho = 0.71$). As the availability of NLP researchers affects datasets' availability, a linguistically diverse group of NLP researchers is required for equitable dataset development.

\paragraph{Availability of Crowdworkers.}
\label{prolific}
We estimate the demographics on the crowdsourcing platform using worker's demographic statistics per their first language listed on the academic research crowdsourcing Prolific,\footnote{\url{https://prolific.co}}
a total of 136,884 workers are available, 70\% of whom speak English as their first language, and the remaining 30\% cover 60 additional languages. 
Figure~\ref{fig:prolific_data_num} shows the relationship between the number of annotators and the number of datasets in our survey. We observe a weak correlation of $\rho = 0.58$ (and $\rho=0.59$ when considering only crowdsourced datasets), possibly because Prolific is not yet widely used in the NLP community except for a handful of datasets~\cite{liu-etal-2021-visually}, and our proxy might miss crowdworkers with proficiency in other languages besides their native one. 
However, no other platforms, including Amazon Mechanical Turk (MTurk)\footnote{\url{https://www.mturk.com/}}, the most widely used crowdsourcing platform according to our survey, provide statistics about annotators. To investigate the potential to gather high-quality multilingual data on English-centric crowdsourcing platforms, we conducted the following  pilot study about multilingual worker availability on MTurk.
\section{Pilot Study on Crowdsourcing Multilingual Data on MTurk}

How easy is it to collect multilingual dataset on popular annotation platform (MTurk) at the moment? 
Crowdsourcing enables large-scale, cost efficient data collection; however, for many languages, the number of language-proficient crowdworkers is limited~\cite{garcia-etal-2021-assessing}.  
We quantify the availability of MTurk workers with proficiency in non-English languages.

We formulated a four-way sentiment analysis task using the Multilingual Amazon Review Corpus~\cite{marc_reviews} and analyzed the annotation quality, cost, and time to finish tasks in English, Spanish, German, French, Japanese and Chinese of crowdworkers. 

In all settings, we asked annotators to translate the same English sentence to assess their actual (rather than professed) language proficiency. We found that many production-level MT systems fail to translate his sentence due to its compositionality. 
Further, we investigated the newly introduced ``language qualification'' in MTurk, which available for only four aforementioned languages and Brazilian Portuguese as of 2022.

For our sentiment analysis task, without the language qualification, the accuracy of human binary classification performance in all non-English languages (55.2\%) was significantly worse than that of English (77\%). Past recommendations, such as constraining location and HIT acceptance rate, are insufficient (as of 2022) to collect high quality data even for languages considered ``easier'' to crowdsource in prior work~\cite{pavlick-etal-2014-language}. 
Language qualification improved performance by up to 40\% and reduced the prevalence of cheating across all languages. 
However, with the language qualification, the data collection process usually took more time and cost (\$1 per assignment). 
More pilot study details are in Appendix~\ref{sec:amt}.
\vspace{-.3em}
\paragraph{Quality Control Using Translation Task.}
We investigate whether crowdworkers relied on automatic machine translation, despite our instruction saying \textbf{not} to use them. We ask native speakers to compare the crowdsourced translation with the translation results from three major translation platforms: Google Translate,\footnote{\url{https://translate.google.com}} Microsoft Bing Translator,\footnote{\url{https://www.bing.com/translator}} and DeepL Translator.\footnote{\url{https://www.deepl.com}} Without language qualification, we identified 33\% of crowdworkers copy-and-pasted automatic translation outputs (with qualification, 7\%). 
This is significantly higher than what \citet{pavlick-etal-2014-language} report (10\%), suggesting more crowdworkers have started to use MT services. 

We found that we could potentially use the translation task to identify good submissions: if we take only the submissions whose translation (1) do not match translation from MT and (2) valid translated judged by native speakers (labeled as either correct or partially correct),\footnote{The details of translation quality study can be found in the appendix.} binary task accuracy rises to 81.0\% from 63.5\%, matching the binary accuracy of English. 

Our pilot study suggests that translation quality can reflect the target task performance if workers who copy from MT systems are filtered, and can be a good proxy for the languages without aforementioned language qualifications. 


\section{Discussion and Suggestions}
\label{discussion}
This work provides the first large-scale meta survey on public multilingual NLP datasets, focusing on the novel aspects under-explored in prior work. 
We found that many languages lack a diverse set of \textbf{manually annotated} datasets and coverage of tasks and input\&label sources. 
Particularly, except for summarization, generation tasks for non-English languages show limited language coverage. 
We conclude this work by presenting concrete suggestions to both the NLP community (Section~\ref{sec:community_suggestions}) and to individual researchers aspiring to create new multilingual NLP datasets (Section~\ref{sec:researcher_suggestions}).

\subsection{Suggestions for the NLP Community}
\label{sec:community_suggestions}


\paragraph{To Foster Language-proficient Researchers and Community Efforts.}
Our analysis shows that the availability of NLP researchers who are fluent in languages highly correlates with the availability of datasets. 
Moreover, monolingual test suites cover only 16 languages, such as Chinese~\cite{xu-etal-2020-clue}, {Indic Languages}~\cite{kakwani-etal-2020-indicnlpsuite}, Polish~\cite{rybak-etal-2020-klej}, Persian~\cite{khashabi-etal-2021-parsinlu}, Russian~\cite{shavrina-etal-2020-russiansuperglue} or Arabic~\cite{seelawi-etal-2021-alue}, where efforts are driven by language-proficient NLP researchers. Organizing these large-scale, inter-organization efforts can be challenging but have profound effects. 
Recent community efforts such as Masakhane\footnote{\url{https://www.masakhane.io/}} spur research for under-resourced languages, resulting in new valuable resources for underrepresented languages (e.g., MasakhaNER; \citealt{10.1162/tacl_a_00416}). 
Developing a directory of language-proficient NLP researchers interested in global collaboration could foster more cooperation. In the long run, globalized NLP education like AFIRM\footnote{\url{https://sigir.org/afirm2020/}} will be necessary. {A directory of potential funding sources to support multlingual data collection can also be helpful. 

\paragraph{On Inclusive Venues.}
The academic publication/conference reviewing system should also reward efforts to develop language-specific resources, without perceiving this as a niche, low-impact effort~\cite{Rogers2021QADE}.
As a community, we should encourage efforts to create and provide region-specific (e.g., Nordic Conference on Computational Linguistics, Pacific Asia Conference on Language, Information and Computation), language-oriented (e.g., Deep Learning for Low-Resource NLP, AfricaNLP, Workshop on Indian Language Data: Resources and Evaluation), and data-oriented (e.g., NeurIPS dataset and benchmark track) venues for introducing multilingual datasets. 

\paragraph{On Multilingual Shared Tasks.}
Several recent shared tasks have driven dataset creation for  both low-resource languages and novel tasks. For example, the WMT 2022 General MT task added four new languages pairs (e.g., Ukrainian), and MIA 2022 Workshop 
released the first annotated open-domain QA data in Tagalog and Tamil~\cite{asai-etal-2022-mia}. 
Similarly, the WMT 2022 Large-scale Machine Translation Evaluation for African Language track\footnote{\url{https://statmt.org/wmt22/large-scale-multilingual-translation-task.html}} presents a data collection track for African languages. 
Large-scale multilingual NLP shared tasks have often focused on major, particularly European languages~\cite{callison-burch-etal-2010-findings,ws-2009-natural-language}, leaving many world languages behind. 
Adapting existing systems to new and low-resource languages poses a challenging and intriguing task as well as substantial  research inquiries. 
The community should continue supporting such efforts and expand evaluation data for diverse target languages.


\subsection{Suggestions for Individual Researchers}
\label{sec:researcher_suggestions}

\paragraph{For Crowdsourcing.}
Our pilot study reveals both the difficulty of crowdsourcing for non-English languages and the high reliance on MT systems on English-centric platform. {To conduct crowdsourcing on MTurk, one can either (1) adding language qualification {newly introduced on MTurk} for the 5 languages available, (2) introducing translation qualification and pruning workers based on their translation quality, and (3) translate original input into English and then crowdsource in English~\cite{xorqa}. }
We also recommend using language-specific crowdsourcing platforms, when available.\footnote{Toloka (\url{https://toloka.yandex.com}) is widely used by Russian language researchers. SelectStar (\url{https://selectstar.ai}) and DeepNatural (\url{https://deepnatural.ai}) are South Korea-based crowdsourcing platforms.} Alternative crowdsourcing platforms like Prolific or freelance platforms, such as CrowdFlower\footnote{\url{https://visit.figure-eight.com/People-Powered-Data-Enrichment_T}} 
or Upwork,\footnote{\url{https://www.upwork.com}} can be explored, though they tend to be more expensive. 

\paragraph{For Translating English Datasets.}
Another option to create a multilingual dataset is to translate datasets in high-resource languages into target languages~\cite{conneau2018xnli,lewis-etal-2020-mlqa}. 
Fortunately, there are many crowdsourced translation services that offer semi-professional translation at cheaper costs {and better availabilities} than translation services provided by trained professional translators. 
In our survey, Gengo\footnote{\url{https://gengo.com}} and One Hour Translation,\footnote{\url{https://onehourtranslation.com}} 
are the most highly used platforms for translation-based multilingual dataset creation. However, {translation artifacts in these datasets remain unclear}. 
Future studies can further quantify the {quality of each translation method} 
in existing translation-based datasets{, the distribution of translation artifacts and mistakes, and the impact of such artifacts on final downstream task performances using our meta-annotations.}
{\paragraph{On Funding Sources.}
\label{sec:funding}
As discussed previously, multilingual dataset creation is often more expensive than English dataset creation.
We summarize funding sources from the paper we surveyed that had over 50 citations. 
For general multilingual research, funding sources mostly consist of national funding agencies, such as the Spanish Ministry of Education and Science, the Catalan Secretary of Linguistic Policy, the Science Foundation Ireland, 
the Irish Research Council, the National Natural Science Foundation of China,  
the Department of Defense (e.g., DARPA, ARL, ARO), the US National Science Foundation, 
and the National Centre for Human Language Technology in the South African Department of Arts and Culture. 
For computational supports, researchers could apply to Google's Tensorflow Research Cloud 
and NVIDIA's academic hardware grant. }

\section{Conclusion}
We present the first large-scale comprehensive survey on characteristics of multilingual datasets,  exposing that the disparity among languages is not only quantitative (i.e., the number of the datasets) but also qualitative (e.g., {\it how} and {\it why} those datasets are created). 
We also discuss building blocks for constructing data resources, from language proficient researchers to crowdworkers. Our MTurk experiments show the challenges of quality and costs of annotating multilingual datasets on MTurk and  suggest several approaches to tackle those challenges.
We conclude our survey with a list of concrete suggestions for researchers interested in constructing language resources.

\section*{Limitations}

\paragraph{On Dataset Documentation. } 
Throughout our survey process, we found inadequate dataset documentation, limiting the coverage of our survey. We suggest that individual researchers provide the input data source and the labeling methodology;  if people were involved in dataset creation, their demographic information should be provided, as well. Such information can help researchers analyze potential bias embedded in the dataset~\cite{bender-friedman-2018-data}. 

\paragraph{Surveyed Dataset Collection Process.}
Despite our best efforts, we do not claim to cover all relevant datasets. Our collection process overlooks datasets that are published at non-ACL venues and not in Hugging Face as well as papers that do not match our search keywords. For instance, we missed multilingual machine reading comprehension datasets~\citep{gupta2020bert,asai2018multilingual} and morphology datasets~\cite{mccarthy-etal-2020-unimorph}. We also found a very low presence of indigenous language datasets. None of 10 indigenous American languages from a recent study~\citep{ebrahimi-etal-2022-americasnli} was represented in our survey. 
That said, we host \url{http://multilingual-dataset-survey.github.io} where researchers can submit their dataset information and periodically update our analysis. Furthermore, we constantly encountered poorly written documentation or unavailable datasets during our annotation processes. During annotation, whenever this paper's dataset  annotators  encountered unclear documentation, they made their best guess to put datasets into predefined categories. If no evidence could be found for the inference, they put ''not mentioned'' as a result. All unclear decision were adjudicated by at least three annotators.

\paragraph{Using Country Names as a Proxy for Languages Spoken.} In Section \ref{sec:correlation}, we attempted to approximate the number of NLP researchers with language proficiency in different languages. To do this, we mapped the names of ACL submission country to the most commonly spoken languages in those countries. We acknowledge that (1) the country of origin of researchers might be different from the country of submission, (2) researchers native language might not be listed in the commonly spoken languages and (3) the mapping might be incomprehensive. 
\paragraph{MTurk Pilot Study.} {Due to our limited data points, although our MTurk study showed that data quality could be improved if the language qualification were applied in the collection process on MTurk, and our previous recommendations do not currently apply, we acknowledge that more research at scale should be done to statistically confirm the conclusion. Furthermore, languages other than the supported 5 languages might still be unsuitable for gathering multilingual data on MTurk}

\section*{Acknowledgements}
We thank Melanie Sclar, Thibault Sellam, and Tobias Rohde for evaluating our mTurk experiment translations. We thank Alisa Liu, Peter West, Zhaofeng Wu, Li Du, Kyle Mahowald and the members in the UW NLP group for their helpful feedback on this work. 
\bibliography{anthology,custom}
\bibliographystyle{acl_natbib}

\appendix
\clearpage
\section*{Appendix}
\label{sec:appendix}

\section{Additional Data Analysis}
\subsection{Task types and input sources.}
We anticipate correlations between the task type and the source of input text. We visualize it in Figure~\ref{fig:task_source_relationship}, and highlight a few findings. 
First, QA (with and without retrieval) datasets are often created using either Wikipedia articles such as \citet{dhoffschmidt-etal-2020-fquad} or curated sources such as exams and scientific papers \cite{vilares-gomez-rodriguez-2019-head}, while {social media such as Twitter} or commercial websites like Amazon.com are mainly used to construct sentiment analysis datasets.
Secondly, summarization dataset mostly derived from news media, MRC and sentence pair classification tasks often involved multiple input sources. For instance, evidence passage can come from existing Wikipedia passage, but the question is crowdsourced~\cite{lewis-etal-2020-mlqa}.

\section{Details of the Meta Analysis}
\subsection{Availability of Unlabeled Text}
\label{sec:input-text-avail}
Unlabeled text can be used for pre-training~\cite{blasi-etal-2022-systematic} or as input sources of the new labeled datasets.
\citet{joshi-etal-2020-state} reports a correlation between the amount of unlabeled data such as Wikipedia articles and the number of datasets on the LDC catalog.\footnote{{\url{https://catalog.ldc.upenn.edu/}}} 
We study the correlation between unlabeled corpora and our surveyed datasets, most of which are not included in licensed LDC. 
As we identify that many datasets use texts beyond Wikipedia (see Table~\ref{tab:input_source}), we instead use the mC4 corpora~\cite{xue-etal-2021-mt5}, a much larger collection of texts in 101 languages drawn from the public Common Crawl web scrape and used for training the mT5 model. Specifically, we use the number of tokens in mC4 to estimate the amount of unlabeled data.
Figure~\ref{fig:mc4} shows a scatter plot where the $x$-axis represents the number of tokens in mC4 and $y$-axis represents the number of labeled datasets available in the languages. 
{The availability of unlabeled text corpora and the number of labeled datasets show a high correlation ($\rho=0.794$).} 
We also analyze the relationship between the number of labeled datasets and the number of Wikipedia articles in Figure~\ref{fig:wikipedia}. Again we observe a high correlation of $\rho = 0.767$. 

\begin{figure}[t!]
\begin{subfigure}[t]{\linewidth}
 \input{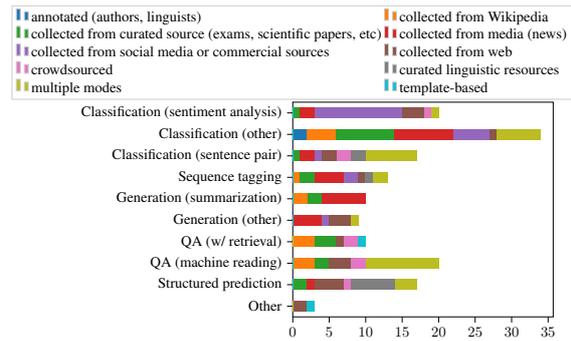}
  \captionsetup{width=\textwidth}
\end{subfigure}%

\vspace{-0.6em}
\caption{The distribution over input source per task type. The size of bar represents the \# of datasets of that task type and input source. \label{fig:task_source_relationship}}
\vspace{-0.6em}
\end{figure} 
\begin{figure}[t!]
\centering
\begin{subfigure}[t]{.5\linewidth}
  \centering
  \includegraphics[width=\textwidth,keepaspectratio]{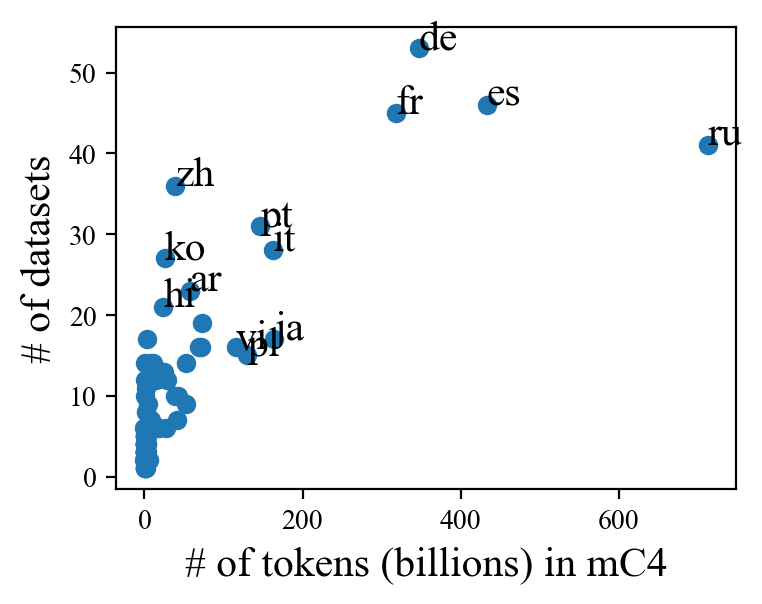}
  \captionsetup{width=0.8\textwidth}
  \caption{The number of tokens (billions) in mC4~\cite{xue-etal-2021-mt5}.}
  \label{fig:mc4}
\end{subfigure}%
\begin{subfigure}[t]{.5\linewidth}
  \centering
  \includegraphics[width=\textwidth,keepaspectratio]{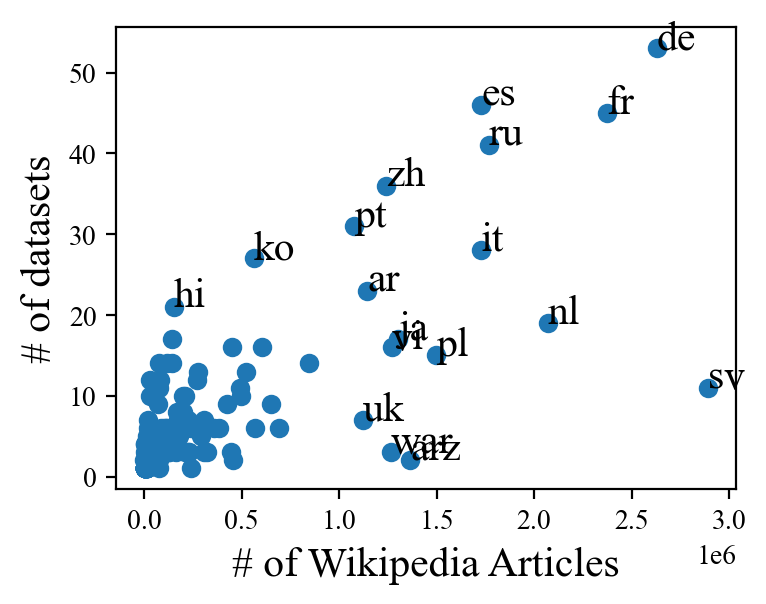}
  \captionsetup{width=0.8\textwidth}
  \caption{The number of Wikipedia articles.}
  \label{fig:wikipedia}
\end{subfigure}%
\vspace{-0.4em}
\caption{Relationships between the number of datasets and number of tokens in mC4 and Wikipedia articles. English is removed. }

\vspace{-0.4em}
\label{fig:appendix-bottleneck}
\end{figure}


\begin{table*}[ht!]
    \small
    \centering
    \begin{tabular}{l||ccccccc}
    \toprule
    Language & {Afrikaans} & {Albanian} & {Amharic} & {Arabic}  & {Armenian} & {Basque} & {Bengali}\\\hline
    \# Workers & 774 & 70 & 25 & 417 & 39 & 27 & 192 \\\bottomrule
    Language  & {Bulgarian} & {Cantonese} & {Catalan} & {\bf Chinese} & {Croatian} & {Czech} & {Danish}\\\hline
    \# Workers  & 168 & 80 & 149 & {\bf 1241} & 101 & 324 & 156\\\bottomrule
    Language & {\bf Dutch} & {\bf English} & {Estonian} & {Farsi} & {Finnish} & {\bf French} & {\bf German}  \\\hline
    \# Workers  &{\bf 1551} & {\bf 93056} & 375 & 127 & 219 & {\bf 1870}  & {\bf 3220}\\
    \bottomrule
    Language & {\bf Greek} & {Gujarati} & {Hebrew} & {Hindi} & {Hungarian} & {Icelandic} & {Indonesian} \\\hline
    \# Workers  & {\bf 1391} & 74 & 387 & 245 & 955 & 25 & 105\\
    \bottomrule
    Language  & {\bf Italian} & {Japanese} & {Khmer} & {Korean} & {Latvian} & {Lithuanian} & {Macedonian}  \\\hline
    \# Workers   & {\bf 3762} & 95 & 27 & 287  & 273 & 133 & 30\\
    \bottomrule
    Language &  {Malay} & {Malayalam} & {Mandarin} & {Nepali} & {Norwegian} & {\bf Polish} & {\bf Portuguese} \\\hline
    \# Workers  & 41 & 64 & 34 & 88 & 129 & {\bf 5609} & {\bf 5948}\\
    \bottomrule
    Language & {Punjabi} & {Romanian} & {Russian}  & {Serbian} & {Slovak} & {Slovenian} & {\bf Spanish} \\\hline
    \# Workers  & 96 & 328 & 715  & 76 & 62  & 377 & {\bf 9060}\\
    \bottomrule
    Language & {Swahili} & {Swedish} & {Tagalog-Filipino} & {Tamil} & {Telugu} & {Thai} & {Turkish}  \\\hline
    \# Workers  & 89  & 377 & 422 & 108 & 67 & 43 & 288 \\\bottomrule
    Language & {Twi} & {Ukrainian} & {Urdu} & {Vietnamese} & {Welsh} \\\hline
    \# Worker & 31 & 52 & 276 & 445 & 89\\\bottomrule
    \bottomrule
    \end{tabular}
    \caption{The number of active crowd-workers in the last 90 days available on Prolific (as of January 6th, 2022) in terms of first language. Note that the number of active workers that has a first language of either Belarusian, Burmese, Dari, Dzongkha, Esperanto, Faroese, Gaelic, Galician, Georgian, Hakka, Inuktitut (Eskimo), Kurdish, Laotian, Lappish, Maltese, Papiamento, Pashto, Scots, Somali, Tajik, Tibetan, Tigrinya, Tongan, Turkmen, or Uzbek is less than 25, and therefore we omit them in the table.}
    \label{table:turker-availability}
\end{table*}

\section{Pilot Study: Investigating the Viability of Crowdsourcing for Six Languages}
\label{sec:amt}
Previous work~\cite{callison-burch-2009-fast,bloodgood-callison-burch-2010-using} studied the feasibility of using crowdsourcing platform {to evaluate} machine translation systems. \citet{pavlick-etal-2014-language} expanded the study to translating 100 languages and recommended several ``best'' languages (high quality results with fast completion speed) to target on MTurk platform. We re-visit the worker availability eight years later, comparing our findings with the previous findings. 

\label{sec:amt-interface}
\begin{figure*}[ht!]
    \centering
    \includegraphics[width=\linewidth]{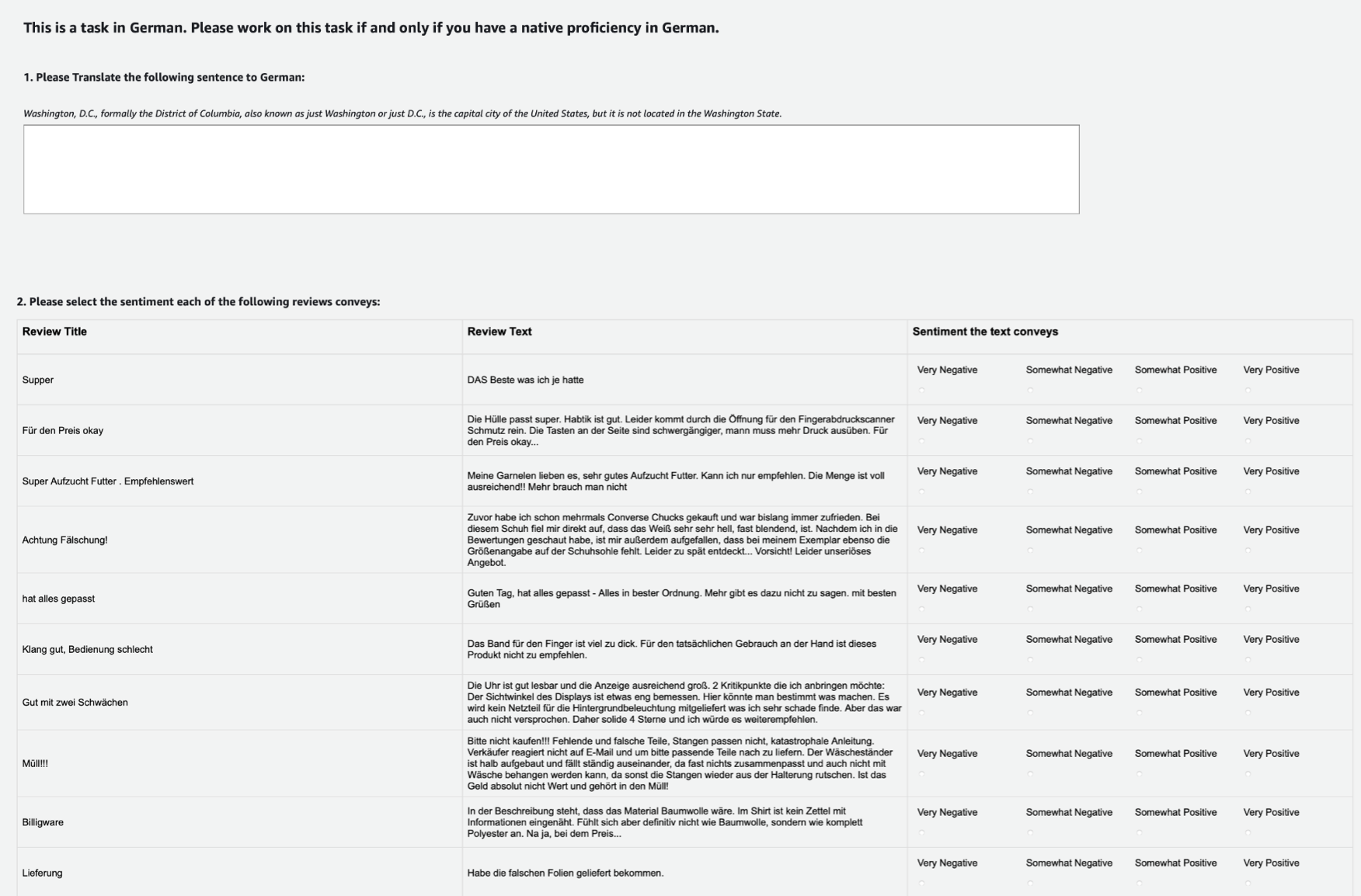}
    \caption{Layout of the tasks on Amazon Mechanical Turk for German}
    \label{fig:amt-layout}
\end{figure*}

\subsection{Task Design}
We design a sentiment analysis task that is trivial for a native speaker but is challenging for someone who does not have native proficiency, {along with a translation task to evaluate workers' true proficiency in the language of interest}. Our sample interface layout is shown in  \ref{fig:amt-layout}. 

\paragraph{Source data and languages.}
We use the Multilingual Amazon Review Corpus (MARC; \citealt{marc_reviews}), which contains 5-way sentiment labels for reviews from Amazon in English, German, French, Spanish, Japanese and Chinese (Mandarin). While all of them are relatively high-resource languages, 
their numbers of the available crowdworkers varies significantly 
(e.g., English has 93K annotators while Japanese only has 95 on Prolific).

\paragraph{Sentiment analysis.}
We extract the reviews with 1 (very negative), 2 (somewhat negative), 4 (somewhat positive) and 5 (very positive) stars, omitting reviews with 3 stars to keep annotation task less ambiguous. We evenly sample 5 reviews of each of four ratings for each respective language. We compute four-way classification accuracy, and binary classification accuracy by merging 1, 2-star ratings as {\it negative} and 4, 5-star ratings as {\it positive}. 

\paragraph{Translation.}
For non-English tasks, as an additional check (for worker's understanding ability and cheating detection), we require crowdworkers to translate a simple yet compositional English sentence from an Wikipedia article,\footnote{\url{https://en.wikipedia.org/wiki/Washington,_D.C.}} \textit{``Washington, D.C., formally the District of Columbia, also known as just Washington or just D.C., is the capital city of the United States, but it is not located in the Washington State.''}
Some widely-used online translation tools give sub-par translations on this sentence because of its compositional structure. We forbid workers to use online translation platforms in the task prompt. 

We collect the gold translation from native speakers as references. In addition, we ask them to rate the collected translations as correct (3), partially correct (2), or incorrect (1).\footnote{A partially correct translation has some minor grammatical errors or lose details, but overall conveys the information.} We also compute the BLEU score against the reference answer. 

\begin{table*}[ht!]
    \small
    \centering
    \begin{tabular}{lp{4.5cm}p{4.5cm}p{4.5cm}}
    \toprule
        lang & Gold tranlsation & w/qual translation samples & w/o qual translation samples \\\midrule
        de & Washington, D.C., formal der District of Columbia, auch bekannt als Washington oder nur D.C., ist die Hauptstadt der Vereinigten Staaten,
        liegt aber nicht im Staat Washington. & Washington, D.C., früher District of Columbia, auch nur Washington oder nur D.C. genannt, ist die Hauptstadt der Vereinigten Staaten, befindet sich jedoch nicht im Bundesstaat Washington. & Washington D.C., formal "District of Columbia" genannt, auch bekannt als "Washington" oder nur "D.C." ist die Hauptstadt der USA, liegt aber nicht im Staat Washington \\
        & & Washington, D.C., früher District of Columbia, auch nur Washington oder nur D.C. genannt, ist die Hauptstadt der Vereinigten Staaten, befindet sich jedoch nicht im Bundesstaat Washington & Washington D.C., formal der "District of Columbia", auch bekannt als "Washington" oder nur "D.C.", ist die Hauptstadt der USA, liegt aber nicht im Staat Washington.\\\hline
        es & Washington, D.C., formalmente el Distrito de Columbia, también conocido simplemente como Washington o como D.C., es la ciudad capital de los Estados Unidos, pero no se encuentra en el Estado de Washington. & Washington, D.C., formalmente el Distrito de Columbia, también conocido simplemente como Washington o simplemente D.C., es la ciudad capital de los Estados Unidos, pero no está ubicada en el estado de Washington. & Washingto, D.c.,formalmente el Distrito de Columbia, tambein conocido simplemente como Washington o simplemente D.C., es la ciudad capital de los estados unidos, pero  no se encuentra en el estado de washington.\\
        & & Washington,D.C.,formalmente el Distrito de Columbia, también conocido simplemente como Washington o simplemente D.C.,es la ciudad capital de los Estados Unidos,pero no se encuentra enel estado de Washington. & Washington, D.C., oficialmente el Distrito de Columbia, también conocido tan solo como Washington, o únicamente D.C., es la ciudad capital de los Estados Unidos, pero no se encuentra en el estado de Washington.\\\hline
        fr & Washington, D.C., officiellement nommée District of Columbia, aussi connue sous le simple nom de Washington ou juste D.C., est la capitale des Etats-Unis, mais elle n'est pas située dans l'Etat de Washington. & Washington, D.C., anciennement le District de Columbia, également connu sous le nom de Washington ou simplement D.C., est la capitale des États-Unis, mais elle n'est pas située dans l'État de Washington. & Washington, D.C, officiellement le District de Columbia, aussi connue juste comme Washington ou juste D.C, est la capitale des Etats-Unis, mais n'est pas située dans l'etat de Washington.\\
        & & Washington, D.C., anciennement le District de Columbia, également connu sous le nom de Washington ou simplement D.C., est la capitale des États-Unis, mais elle n'est pas située dans l'État de Washington. 
        & Washington D.C, officiellement The District of Columbia, également connue sous le nom de Washington, ou simplement D.C, est la capitale des États-Unis, mais n'est pas située dans l'état de Washington.\\\hline
        ja & \begin{CJK}{UTF8}{min}ワシントンDCは公式にはコロンビア特別区、もしくは単にワシントンおよびDCと呼ばれ、アメリカ合衆国の首都ではあるがワシントン州に位置しているわけではない。\end{CJK}& \begin{CJK}{UTF8}{min}ワシントンD.C.、正式にはコロンビア特別区、別名ワシントンD.C.は米国の首都ですが、ワシントン州にはありません。  \end{CJK}& ---\\\hline
        zh-cn & \begin{CJK}{UTF8}{gbsn}
        华盛顿特区，正式名称为哥伦比亚特区，也被称为华盛顿或D.C.，是美国的首都，但它并不位于华盛顿州内。
        \end{CJK} & \begin{CJK}{UTF8}{gbsn} 国会图书馆，华盛顿特区。新的联邦领土被命名为哥伦比亚特区，以纪念探险家克里斯托弗·哥伦布，新的联邦城市以乔治·华盛顿的名字命名\end{CJK}& \begin{CJK}{UTF8}{gbsn}华盛顿DC，理论上叫哥伦比亚特区，也被称作华盛顿或者DC，是美国的首都，但是不位于华盛顿州\end{CJK}\\
        zh-tw & \begin{CJK}{UTF8}{bsmi}華盛頓特區，正式名稱為哥倫比亞特區，也被稱為華盛頓或D.C.，是美國的首都，但它並不位於華盛頓州內。 \end{CJK}& \begin{CJK}{UTF8}{bsmi}華盛頓特區，正式名稱為哥倫比亞特區，也稱為華盛頓或華盛頓特區，是美國的首都，但並不位於華盛頓州。\end{CJK} & --- \\\bottomrule
    \end{tabular}
    \caption{Sample translation results, We provide gold translation and select samples for both simplified Chinese and traditional Chinese to display.}
    \label{table:amt-translation}
\end{table*}
We present some sample translation results as well as the gold translation provided by human native speaker annotators in Table~\ref{table:amt-translation}.
We found that the most easily made mistake is the misreading of the word ``formally'' to ``formerly'', which also exists from the google translated German sentence, and it also happened when we initially ask native speakers to translate the sentence.
\subsection{Task Setting}
\paragraph{Worker qualification.} 

We use following qualifications provided by MTurk: (i) workers must have at least a 95\% acceptance rate, (ii) workers must be in the US or Top 5 countries with the largest population speaking for each
    language from WorldData.info,\footnote{\url{https://www.worlddata.info}} and (iii) for Spanish, German, French and Chinese with ``Language Fluency (Basic)'' premium qualifications available, we {collect labels on the same data with and without this qualification}. Note that as of 2022, it's only available for {the four aforementioned languages} and Brazilian Portuguese with an additional \$1 per assignment.
    

\paragraph{Task statistics.}
We collect 20 sentiment annotations per languages along with one translation example, each of which accepts up to 10 unique MTurkers' annotations, resulting in 200 annotations for all six languages ({\bf without language qualification}). 
In Spanish, Germany, French and Chinese, we release the same HITs with the language proficiency requirements, resulting in additional 200 annotations ({\bf with language qualification}). We aimed at an hourly pay of \$12 for this task.

\subsection{Evaluation and Analysis}
 
Table \ref{tab:amt-meta} shows the time elapsed, the number of annotations collected, the four-way and binary accuracy for the sentiment analysis task, the average human evaluation score and sacreBLEU~\cite{post-2018-call} score for translation task for each of the language experiments with and without the qualification. 

\paragraph{Time elapsed to collect annotations.}
Without language qualification, all annotations finished in a single day, with English being the fastest, and Chinese being the slowest (0.8 v.s. 6.9 hours). 
When language proficiency criteria is added, the tasks expired after four days without gathering all annotations, with Spanish being the most available and Chinese being the least available (140 v.s. 30 annotations). Conversely, their language qualification may overly shrinks the worker pool; 
according to a web forum among crowdworkers,\footnote{\url{https://turkerview.com/qualifeye/}} paths to acquiring this language qualification is unclear. We assume that many workers with proficiency in the target language might not obtain the qualification. 

\paragraph{Annotation quality on sentiment analysis task.}
We evaluate the binary and 4-way classification accuracy. Random baseline yield 50\% and 25\% accuracy, respectively. 
Although \citet{pavlick-etal-2014-language} recommends French, German, and Spanish to be some of the best target language on MTurk, our results were unsatisfactory on these languages; without language qualification, the classification performance for all languages is significantly worse than English (77\%), indicating that only by constraining location and HIT acceptance is insufficient as of 2022. {Language qualification improves performance across languages (e.g., 46.5\% and 30.6\% 4-way accuracy improvements in Germany and Spanish, respectively). }
\begin{table*}[ht!]
\footnotesize
    \centering
    \begin{tabular}{{p{1cm}p{1.65cm}p{1.5cm}p{1.3cm}p{1.8cm}p{1.8cm}p{1.8cm}p{1.8cm}}}
    \toprule
         language &\# annotations  & time (hour) & \# matched & 4-way acc(\%) & binary acc(\%) & human score & sacreBLEU \\\midrule
         English  & 200 / -- & 0.83 / -- & -- & 42.0 / -- & \textbf{77.0} / -- & -- & -- \\
         Spanish & 200 / 140 & 1.43 / 96 & 6 / 0 & 28.0 / 58.6 & 51.5 / 95.0 & \textbf{1.80} / 2.36 & 39.09 / \textbf{63.09} \\
         German & 200 / 40 & 1.30 / 96 & 11 / 0 &  23.5 / \textbf{70.0}& \textit{48.5} / 95.0 & 1.70 / \textbf{2.50} & \textbf{39.13} / 57.33 \\
         French & 200 / 60 & 1.25 / 96 & 8 / 2& \textit{23.0} / 50.0 &  57.0 / \textbf{100} & 1.55 / 2.33 & 33.30 / 49.33\\
         Japanese & 200 / -- & 3.78 / -- & 10 / -- & \textbf{44.0} / -- & 65.5 / -- & 1.74 / -- & 10.20 / -- \\
         Chinese & 200 / 30 & 6.98 / 96 & 5 / 0 & 29.5 / 56.7 & 53.5 / 80.0 & 1.60 / 2.33 & 32.03 / 42.63 \\\bottomrule
    \end{tabular}
    \caption{{Results from MTurk pilot study for data collection on six languages. Each cell reports the results from without / with language qualification. We report the number collected annotations, elapsed time to finish (max 96 hours which is when the HIT expires), the number of annotators whose translation matched the output from MT systems, average binary classification accuracy and 4-way classification accuracy of sentiment analysis task, average human evaluation score and sacreBLEU score for translation task. }}
    \label{tab:amt-meta}
\end{table*}
\paragraph{Annotation quality on translation task.}
Table~\ref{tab:amt-meta} shows that the average sacreBLEU score without language qualification is significantly lower than the one with qualification for all the languages. 
As shown in Table~\ref{tab:amt-meta}, human rating with language qualification is higher (e.g., 2.5 v.s. 1.7 in German). The human evaluation correlates with sacreBLEU metric ($\rho = 0.79$). Without language qualification, it is non trivial to collect high-quality translation data from workers proficient in the target language.

\end{document}